\pdfoutput=1

\documentclass[11pt,a4paper]{article}
\usepackage[hyperref]{acl2020}
\usepackage{times}
\usepackage{latexsym}

\usepackage{graphicx}
\usepackage{multirow}
\usepackage{multicol}

\usepackage{microtype}

\aclfinalcopy

\usepackage{textcomp}

\title{Neutron: An Implementation of the Transformer Translation Model\\ and its Variants}

\author{Hongfei Xu \\
  Saarland University\\
  DFKI \\
  Saarbruecken, 66123 \\
  \texttt{hfxunlp@foxmail.com} \\\And
  Qiuhui Liu \\
  China Mobile Online Services\\
  Henan, 450001 \\
  \texttt{liuqiuhui@cmos.chinamobile.com} \\}

\date{}

\begin{document}
\maketitle
\begin{abstract}
The Transformer translation model is easier to parallelize and provides better performance compared to recurrent seq2seq models, which makes it popular among industry and research community. We implement the Neutron \footnote{We open source our implementation at \url{https://github.com/anoidgit/Transformer}.} in this work, including the Transformer model and its several variants from most recent researches. It is highly optimized, easy to modify and provides comparable performance with interesting features while keeping readability.
\end{abstract}

\section{Introduction}

\newcite{vaswani2017attention} proposed Transformer architecture which contains only attention mechanism, standard feed-forward neural network with residual connection, dropout, layer normalization and label smoothing loss for its training. The Transformer parellelizes better and outperforms previous seq2seq models in many cases. It is widely applied in the industry and attracts wide attention from researchers, as a result, many enhanced architectures have been further proposed.

Our implementation first supports those popular features provided in most Machine Translation (MT) libraries, including beam search, ensemble, length penalty and averaging of models.

In addition, we implement some variants of the Transformer from recent researches along with the standard implementation of Transformer in the Neutron, such as: the Average Attention (to accelerate the decoding of the Transformer) \cite{zhang2018accelerating}, the Hierarchical Layer Aggregation \cite{dou2018exploiting}, Recurrent decoder \cite{chen2018best}, Sentential Context \cite{wang2019exploiting} for improving the MT quality, the Transparent Attention \cite{bapna2018training} for the convergence of deep encoders and the document-level Transformer \cite{zhang2018improving}. Neutron also supports more efficient training scheduler (dynamic sampling and review mechanism) proposed by \newcite{wang2018dynamic}, and the other advanced features.

We will introduce features of the Neutron in the next section, its design in the third section, performance in the fourth section, followed by related work and conclusion.

\section{Features}

Neutron supports a wide range of features for research purpose.

\subsection{Basic Support.}

\paragraph{Basic Features.} We support a wide range of fundamental approaches which are usually used in MT, including: beam search of batches, length penalty in beam search and ensemble of models.

\paragraph{Gradient Accumulation.} The batch size of the Transformer affects its optimization and performance, we implement gradient accumulation which accumulates gradients of smaller batches to achieve theoretically unlimited batch size for the Transformer.

\paragraph{Multi-GPU Parallelization.} The default multi-GPU parallelization model of the pyTorch \cite{adam2019pytorch} automatically synchronizes parameters during the forward pass and collects gradients after backpropagation. However, when working with the gradient accumulation, this results in redundant communication between GPUs, and we provide a new implementation of the parallelization model which requires explicit calls to distribute parameters and to accumulate gradients across GPUs but avoids the redundant communication and significantly accelerates the multi-GPU case. Our multi-GPU parallelization model also supports multi-GPU decoding which are normally not implemented in the other libraries.

\paragraph{Label Smoothing Loss.} We support the label smoothing loss \cite{Szegedy2016Rethinking} applied in the Transformer which optimizes the KL-divergence rather than the perplexity.

\subsection{Models.}

\paragraph{Two Computation Orders.} The official implementation of the Transformer \cite{tensor2tensor} uses a different computation order than the proposed Transformer \cite{vaswani2017attention}, which leads to differences in the performance and the convergence \cite{xu2019deep}, we supported both computation orders in the Neutron.

\paragraph{Average Attention.} \newcite{zhang2018accelerating} propose to use the average attention layer instead of the self attention network in the decoder to accelerate the decoding, and we support their approach.

\paragraph{Transparent Attention.} We implement the Transparent Attention mechanism \cite{bapna2018training} which ensures the convergence of deep Transformer encoders.

\paragraph{Hierarchical Layer Aggregation} \newcite{dou2018exploiting} propose to hierarchically aggregate all layers instead of using only outputs of the last layer, we implement the model which leads to the best performance in their paper.

\paragraph{RNMT Decoder.} We support the recurrent decoder proposed by \newcite{chen2018best} which may lead to better performance than the Transformer decoder.

\paragraph{Sentential Context.} \newcite{wang2019exploiting} propose a model which exploits sentential context in the Transformer, we implement the model which achieves the best performance in their paper.

\paragraph{Document-level Transformer.} \newcite{zhang2018improving} propose a document-level Transformer model to utilize inter-sentence contexts, we implement their approach as a baseline for context-aware NMT.

\subsection{Advanced Features.}

\paragraph{Lipschitz Constrained Parameter Initialization.} \newcite{xu2019deep} propose the Lipschitz Constrained Parameter Initialization to ensure the convergence of deep Transformers which also leads to slight but consistent improvements in BLEU. It is implemented as the default parameter initialization approach.

\paragraph{Reducing Optimization Difficulty.} We propose that the biases of linear transformations which project the multi-head attention results and the hidden activation of position-wise feed-forward networks to residual connections are redundant, as the layer normalization \cite{ba2016layer} adds the bias for the input of the next layer. We suggest that removing redundant biases can reduce the computation costs with small acceleration, and may ease the optimization of models.

\paragraph{Dynamic Sentence Sampling.} \newcite{wang2018dynamic} propose to dynamically sample data which reduces more losses during the training, and we implement their approach in the project.

\paragraph{Activation Functions.} The Transformer uses the ReLU activation function, but we also provide the other activation functions, like GeLU and Swish \cite{ramach2017searching}.

\paragraph{Optimizers.} In addition to the Adam optimizer \cite{Kingma15Adam} used by the Transformer, we also support the other optimizers like: RAdam \cite{liu2019variance}, Lookahead \cite{Zhang2019Lookahead} and their combination.

\subsection{Data Cleaning.}

Normally, parallel data for training an MT system are not collected directly from translators at sentence level. Some of the corpus are crawled from the internet, and automatic sentence alignment tools are applied to extract sentence-level translation pairs. As a result, there might be some wrong translations in the parallel data, and we provide some tools to clean data sets. Removing those dirty data will reduce the size of data and the corresponding vocabulary size, and normally leads to faster training with better quality in practice.

\subsubsection{Max Keeper.}

Parallel data are usually combination of several corpus, and those corpus may contain some common sentence pairs, which will be redundant after concatenation. In addition to that, alignment tools may wrongly align a same source sentence into several translations in different contexts, especially for those short sentences. We collect all sentences and their translations, and only save those translations with highest frequency. We also replace potential repeated blanks or tabular into a single blank during cleaning to normalize the data.

\subsubsection{Cleaning with Vocabulary.}

There are some sentence pairs which are meaningless or even not belonging to language pairs researching on. Vocabulary based cleaning is supported for this case. It first collects the vocabulary and counts frequencies of tokens on training set, and then filter training set with a hyper parameter named \emph{vratio}, \emph{vratio} tokens of full vocabulary with least frequencies will be regarded as rare words, and if the percentage of rared tokens in a sentence is higher than \emph{1.0 - vratio}, the sentence is unlikely to be part of the language pair, and will be removed.

\subsubsection{Cleaning with Length Ratios.}

Length ratio is likely to be used in MT data processing, since there are some wrongly aligned sentence pairs in the training data which have abnormally large length ratios. This work provides enhanced support with sub-word units \cite{sennrich2016neural}, though cleaning on only tokenized text is also supported.

Assume a sentence contains \emph{nsub} tokens after BPE processing, \emph{nsubadd} tokens are additionally produced by BPE seperation, \emph{nsep} tokens of original tokenized sentence which has \emph{ntok} tokens are segmented by BPE. Following ratios are defined at monolingual level:

\begin{equation}
    cratio = nsubadd / nsub
\end{equation}
\begin{equation}
    bratio = nsub / ntok
\end{equation}
\begin{equation}
    sratio = nsep / ntok
\end{equation}
Assume a source sentence contains \emph{nsubsrc} sub-word tokens and \emph{nsrc} tokens before applying BPE, \emph{nsubtgt} and \emph{ntgt} correspondingly for its translation, we define two bilingual ratios:
\begin{equation}
    uratio = \frac{{\max (nsubsrc,nsubtgt)}}{{\min (nsubsrc,nsubtgt)}}
\end{equation}
\begin{equation}
    oratio = \frac{{\max (nsrc,ntgt)}}{{\min (nsrc,ntgt)}}
\end{equation}

The reason why we cleaning the training set with ratios related to sub-word units is that those rare words in dirty sentence pairs are likely to be segmented into many sub-word units, which would significantly increase those ratios.

We provide tools to calculate above ratios with validation set, which are good and safe choices for data cleaning.

\subsection{Additional Tools.}

\paragraph{Averaging Models.} The Transformer averages several checkpoints before evaluating, and we support this function which loads parameters of several checkpoints and save averaged parameters to a new model file.

\paragraph{Ranking.} A ranking tool is provided to rank data sets with a pre-trained model, per-token loss will be calculated efficiently. This tool can be employed for data cleaning, domain adapted data selection or in the evaluation of linguistic phenomena.

\paragraph{Web Server.} We provide a simple translation web server with REST API support besides the translating scripts, which we think may be helpful for integrating trained models into the other MT based applications.

\paragraph{Conversion to C Libraries.} Neutron has a converting tool based on Cython\footnote{https://cython.org/} which converts python implementations of core modules and functions into C codes and compile them into loadable C libraries, which may bring little additional performance and make it easier to put the Neutron into practice.

\paragraph{Forbidden Indexes for the Shared Vocabulary.} In practice scenario, there might be some tokens which only appear in the source side when a shared vocabulary is adopted, and these tokens which will never appear in the target side will still get a small smoothing probability in the loss function. We provide a tool to extract those indexes and save it into a file which can be loaded to prevent the effects of those tokens on the decoder and the label smoothing loss.

\begin{table*}[t]
  \centering
    \begin{tabular}{|r|l|r|r|}
    \hline
          & BLEU & Training Speed & Decoding Speed \\
    \hline
        \newcite{vaswani2017attention}  & 27.3  & \multicolumn{2}{|r|}{} \\
    \hline
         Neutron & 28.07 & 21562.98 & 68.25 \\
    \hline
    \end{tabular}
  \caption{Performance and Speed. Training speed and decoding speed are measured by the number of target tokens per second and the number of sentences per second.}
  \label{tab:perf}
\end{table*}

\section{Design}

In this section, we will introduce the design of the Neutron.

\paragraph{Scripts.} We provide scripts for processing training data and testing under ``scripts/'', and the training and decoding scripts are ``train.py'' and ``predict.py''.

\paragraph{Basic Modules.} We implement basic modules under ``modules/'', including the multi-head attention network, the positional embedding, the position-wise feed-forward network, the average attention network, etc.

\paragraph{Loss.} We support the label smoothing loss with ``loss.py''.

\paragraph{Learning Rate Scheduler.} We implement the learning rate of the Transformer \cite{vaswani2017attention} in ``lrsch.py''.

\paragraph{Parallelization.} Our multi-GPU parallelization model is implemented under ``parallel/''.

\paragraph{Supporting Functions.} We implement basic functions for training, decoding and data processing such as: freezing / unfreezing parameters of models, padding list of tensors to same size on assigned dimension under ``utils/''.

\paragraph{Transformer and its Variants} We put the implementation of Transformer models under ``Transformer/''.

\paragraph{Optimizers.} The implementation of optional optimizers can be found under ``optm/''.

\paragraph{Tools.} All tools supporting data processing, translating, etc. are implemented under ``tools''.

\section{Performance}

To compare with \newcite{vaswani2017attention}, we tested our implementation on the WMT 14 English to German news translation task following the settings of \newcite{vaswani2017attention}.

We applied joint Byte-Pair Encoding (BPE) \cite{sennrich2016neural} with $32k$ merge operations and $8$ as the vocabulary threshold for the BPE to address the unknown word issue. We only kept sentences with a maximum of $256$ sub-word tokens for training. The training set was randomly shuffled in every training epoch. The concatenation of newstest 2012 and newstest 2013 was used for validation and newstest 2014 as the test set.

The number of warm-up steps was set to $8k$ \footnote{\url{https://github.com/tensorflow/tensor2tensor/blob/v1.15.4/tensor2tensor/models/transformer.py\#L1818}.}, and each training batch contained at least $25k$ target tokens. We trained the model on $2$ GTX 1080 Ti GPUs, and performed decoding on $1$ of them. We used a dropout of $0.1$. We used the Transformer Base setting \cite{vaswani2017attention}, and the model was trained for $100k$ training steps. We employed a label smoothing \cite{Szegedy2016Rethinking} value of $0.1$. We used the Adam optimizer \cite{Kingma15Adam} with $0.9$, $0.98$ and $10^{-9}$ as $\beta_{1}$, $\beta_{2}$ and $\epsilon$.

We used a beam size of $4$ without length penalty for decoding, and evaluated tokenized case-sensitive BLEU \footnote{\url{https://github.com/moses-smt/mosesdecoder/blob/master/scripts/generic/multi-bleu.perl}.} with the averaged model of the last $5$ checkpoints saved with an interval of $1,500$ training steps \cite{vaswani2017attention}. Results are shown in Table \ref{tab:perf}.

Table \ref{tab:perf} shows that our Neutron implementation of the Transformer surpasses \newcite{vaswani2017attention}, and we suggest our work is helpful for establishing a higher baseline.

\section{Related work}

\newcite{sutskever2014sequence,bahdanau2014neural,luong2015effective,gehring2017convolutional,vaswani2017attention} and many other researchers proposed various kinds of neural machine translation (NMT) models, corresponding to these work, there are many implementations open sourced by researchers like: \newcite{klein2017opennmt,tensor2tensor,hieber2017sockeye,zhang2017thumt}. These implementations greatly help NMT researches and productive applications. Their work also provides valuable experience for us to implement the Neutron.

\section{Conclusion}

We focus on the Transformer and its variants for NMT, implement the Neutron based on the pyTorch which can achieve competitive performance and provides several additional features for NMT. We introduce its features and performance in this paper.

\section*{Acknowledgments}

Hongfei XU acknowledges the support of China Scholarship Council ([2018]3101, 201807040056).

\bibliography{acl2020}
\bibliographystyle{acl_natbib}
\end{document}